\documentclass[lettersize,journal]{IEEEtran}
\usepackage{amsmath,amsfonts}
\usepackage{algorithmic}
\usepackage{algorithm,mathtools,multirow}
\usepackage{array}
\usepackage[caption=false,font=normalsize,labelfont=sf,textfont=sf]{subfig}
\usepackage{textcomp}
\usepackage{stfloats}
\usepackage{hyperref}
\usepackage{graphicx}
\usepackage{color}
\hypersetup{
    colorlinks=true,
    linkcolor=blue,
    citecolor=blue,
    urlcolor=blue
}
\usepackage{url}
\usepackage{verbatim}
\usepackage{cite}
\usepackage[normalem]{ulem}

\usepackage{orcidlink}
\begin{document}

\title{A Coupled Physics-Informed Neural Network for Greenhouse Climate State Reconstruction and Parameter Identification under Sparse Sensor Measurements}

\author{
Sani Biswas\raisebox{0.45ex}{\orcidlink{0009-0000-8888-5293}}, Khursheed J. Ansari{\orcidlink{0000-0003-4564-6211}}, and Md. Nasim Akhtar{\orcidlink{0000-0002-1509-0403}}
        % <-this % stops a space
\thanks{Manuscript submitted \today. This work acknowledges partial funding from Centro de Modelamiento Matem\'atico (CMM) FB210005 BASAL fund for centers of excellence  from ANID-Chile, University of Chile. \textit{(Corresponding author: Sani Biswas)}}% <-this % stops a space

\thanks{Sani Biswas is with  the Centro de Modelamiento Matem\'atico (CMM), University of Chile, 8370456,  Santiago, Chile (email: sani.dumkal@gmail.com).}

\thanks{Khursheed J. Ansari is with the Department of Mathematics, College of Science, King Khalid University, 61413, Abha, Saudi Arabia (email:  kansari@kku.edu.sa, ansari.jkhursheed@gmail.com).}
\thanks{Md. Nasim Akhtar is with Department of Mathematics, Presidency University, 700073, Kolkata, India (email:  nasim.maths@presiuniv.ac.in).
}
}

%\markboth{IEEE TRANSACTIONS ON AUTOMATION SCIENCE AND ENGINEERING,~Vol.~XX, No.~XX, AUGUST~2026}%

%\IEEEpubid{0000--0000/00\$00.00~\copyright~2026 IEEE}

\maketitle

\begin{abstract}
Accurate reconstruction of greenhouse climate variables from sparse sensor measurements is essential for intelligent environmental monitoring, automated climate control, and precision agriculture. In practical greenhouse operation, sensor failures, communication interruptions, calibration drift, and measurement noise frequently result in incomplete observations, making reliable estimation of indoor temperature and relative humidity a challenging inverse problem. This paper presents a coupled physics-informed neural network (PINN) for the simultaneous reconstruction of greenhouse temperature and relative humidity together with the identification of unknown physical parameters governing a reduced greenhouse climate model. The proposed framework integrates measurement data with coupled energy- and moisture-balance equations and initial-condition constraints, enabling greenhouse climate state estimation and physical parameter identification within a unified learning framework. The proposed methodology is evaluated using real greenhouse measurements under two complementary validation protocols: random interpolation from sparse observations (Experiment~A) and chronological temporal extrapolation over an unseen future time interval (Experiment~B). To assess the contribution of the embedded physical constraints, the proposed PINN is systematically compared with a fully connected neural network, a long short-term memory (LSTM) network, and a gated recurrent unit (GRU) network. Under the interpolation setting, the proposed PINN achieves the highest temperature reconstruction accuracy with an RMSE of $0.4495\,^{\circ}\mathrm{C}$ and an $R^2$ value of $0.9636$, while simultaneously identifying physically interpretable model parameters. Furthermore, the complementary evaluation protocols demonstrate that interpolation and temporal extrapolation assess fundamentally different aspects of model performance and provide a more comprehensive evaluation of greenhouse climate reconstruction. The proposed framework provides a practical foundation for intelligent greenhouse monitoring, virtual sensing, digital twins, and next-generation automated greenhouse climate management.
\end{abstract}
\hfill
\section{Introduction}
\IEEEPARstart{G}{reenhouse} cultivation has become a cornerstone of modern controlled-environment agriculture, enabling year-round crop production through precise regulation of environmental conditions while improving crop yield, resource efficiency, and product quality. Recent advances in sensing technologies, wireless communication, artificial intelligence, and automation have accelerated the development of intelligent greenhouse systems capable of continuously monitoring and regulating key environmental variables, including temperature, relative humidity, ventilation, irrigation, and lighting \cite{VanStraten2010,Shamshiri2018,Chen2025, Mowla2023}. These technological advances are transforming conventional greenhouse operation into data-driven intelligent systems that support sustainable agricultural production and autonomous climate management.

Among the environmental variables governing greenhouse operation, indoor temperature and relative humidity are the two most influential factors because they directly affect plant transpiration, photosynthesis, nutrient uptake, disease development, and greenhouse energy consumption. 
Accurate estimation of these variables is therefore fundamental to intelligent greenhouse monitoring, automated climate regulation, and predictive decision-making in modern smart agriculture~\cite{Shamshiri2018,Friha2021}.
In practice, however, sensor failures, communication interruptions, calibration drift, maintenance activities, and measurement noise frequently result in incomplete observations, making reliable estimation of greenhouse climate states a challenging inverse problem. Consequently, robust computational models capable of reconstructing missing climate information from sparse measurements are of considerable practical importance.

Mechanistic greenhouse models are traditionally derived from coupled energy- and moisture-balance equations that describe the dominant heat- and mass-transfer processes inside the greenhouse \cite{Bennis2008,VanStraten2010}. Owing to their strong physical interpretability, these models have been widely employed for greenhouse simulation, optimization, and climate control. Their predictive capability, however, depends heavily on accurate knowledge of greenhouse geometry, crop characteristics, ventilation behaviour, material properties, and numerous physical parameters that are often difficult to determine or may vary over time \cite{Chen2025}. Consequently, purely physics-based models may experience reduced predictive accuracy when observations are incomplete or when the underlying parameters are uncertain.

Driven by recent advances in artificial intelligence, purely data-driven models have become increasingly popular for greenhouse climate prediction. Artificial neural networks, long short-term memory (LSTM) networks, gated recurrent unit (GRU) networks, and related deep-learning approaches can learn complex nonlinear relationships directly from historical observations without requiring explicit physical modelling \cite{Singh2017,Yang2023,Yu2025,Li2026}. Despite their flexibility, these methods primarily capture statistical correlations in the available data. Consequently, their predictive performance may deteriorate when observations are sparse, noisy, or collected under operating conditions that differ from those represented during training. Furthermore, because the governing physical laws are not explicitly incorporated into the learning process, the resulting predictions often lack physical interpretability.

Physics-informed neural networks (PINNs) provide an alternative learning paradigm by integrating observational data with governing physical equations within a unified optimization framework \cite{Raissi2019,Karniadakis2021,Cuomo2022}. Rather than relying exclusively on measurement data, PINNs incorporate the residuals of the governing differential equations into the objective function, encouraging the learned solution to satisfy both the observations and the underlying physical principles. This combination of data fidelity and physical consistency has demonstrated considerable promise in scientific machine learning, particularly for inverse problems, parameter estimation, and state reconstruction under limited observational data.

Although PINNs have attracted increasing attention across a wide range of scientific and engineering applications, their use in greenhouse climate modelling remains relatively limited. Existing studies have primarily focused on greenhouse climate prediction, ventilation modelling, or hybrid physics-guided learning frameworks \cite{Choi2026,Liu2025}. Comparatively little attention has been devoted to the simultaneous reconstruction of coupled greenhouse temperature and humidity together with identification of the underlying physical parameters from sparse observations. Moreover, most existing studies evaluate their methods using only a single validation strategy, making it difficult to distinguish interpolation capability from genuine predictive generalization.

Motivated by these observations, this paper proposes a coupled physics-informed neural network (PINN) for the simultaneous reconstruction of indoor temperature and relative humidity together with the identification of unknown physical parameters governing a reduced greenhouse climate model. The proposed framework embeds coupled energy- and moisture-balance equations directly into the neural-network training process, enabling physical consistency and observational fidelity to be enforced simultaneously. Unlike conventional supervised learning approaches, the unknown physical coefficients are treated as trainable variables and are estimated jointly with the latent greenhouse climate states.

The proposed framework is evaluated using real greenhouse measurements under two complementary experimental protocols. Experiment~A investigates climate reconstruction from randomly sampled observations, representing an interpolation problem within the observed operating regime. Experiment~B employs a chronological train--test partition and evaluates the capability of the proposed model to predict greenhouse climate over a previously unseen future time interval, thereby providing a substantially more challenging temporal extrapolation task. To assess the contribution of the embedded physical constraints, the proposed PINN is systematically compared with a fully connected neural network, an LSTM network, and a GRU network trained under identical experimental settings.

The main contributions of this work are summarized as follows.

\begin{itemize}

\item We develop a coupled PINN for the simultaneous reconstruction of greenhouse temperature and relative humidity from sparse sensor measurements.

\item We formulate greenhouse climate reconstruction and physical parameter identification as a unified inverse problem by treating the unknown physical coefficients as trainable variables within the PINN framework.

\item We perform a comprehensive comparison between the proposed PINN and representative data-driven neural models, including a fully connected neural network, LSTM, and GRU architectures.

\item We evaluate the proposed framework under two complementary validation protocols, namely random interpolation and chronological temporal extrapolation, providing a comprehensive assessment of both reconstruction accuracy and predictive generalization.

\item We demonstrate that embedding physical knowledge into deep learning enables physically interpretable parameter estimation while providing a robust framework for greenhouse climate reconstruction under sparse sensing conditions.

\end{itemize}

\subsection{Related Work} 
Greenhouse climate modelling has been extensively investigated over the past several decades, leading to a diverse range of modelling approaches that can broadly be categorized into mechanistic models, purely data-driven learning methods, and hybrid physics-guided approaches. Each paradigm offers distinct advantages but also exhibits inherent limitations, motivating the development of more robust and physically consistent modelling frameworks.

Mechanistic greenhouse models are generally derived from coupled energy and moisture balance equations that describe heat transfer, ventilation, radiation exchange, evapotranspiration, and mass transport within the greenhouse environment \cite{Bennis2008,VanStraten2010}. Because of their strong physical interpretability, these models have been widely used for greenhouse simulation, climate control, and operational optimization. However, their practical performance depends critically on accurate knowledge of greenhouse geometry, crop characteristics, environmental conditions, and numerous physical parameters that are often difficult to determine or may vary significantly over time. However, parameter uncertainty and incomplete observations can substantially reduce model accuracy when these models are applied to real greenhouse systems \cite{Chen2025}.

Recent advances in artificial intelligence have stimulated extensive research on purely data-driven greenhouse climate prediction. Artificial neural networks (ANNs), support vector machines, ensemble learning methods, recurrent neural networks, LSTM networks, and GRU architectures have all demonstrated promising predictive capabilities for modelling nonlinear greenhouse dynamics \cite{Singh2017,Yang2023,Yu2025,Li2026}. These approaches avoid explicit physical modelling and are capable of learning complex nonlinear relationships directly from historical measurements. Nevertheless, because they rely primarily on statistical correlations, their predictive performance often deteriorates when training data are sparse, noisy, or collected under operating conditions that differ from those represented during model training. In addition, the resulting models generally provide limited physical interpretability and offer little insight into the underlying environmental processes governing greenhouse climate evolution.

PINNs, introduced by Raissi \textit{et al.} \cite{Raissi2019}, have emerged as a powerful framework for integrating physical knowledge with deep learning. By embedding governing differential equations into the optimization process, PINNs enforce physical consistency while simultaneously fitting observational data. This learning paradigm has demonstrated remarkable success across a wide range of scientific and engineering applications, including fluid mechanics, heat transfer, groundwater flow, inverse problems, and parameter estimation \cite{Karniadakis2021,Cuomo2022}. Compared with conventional neural networks, PINNs typically require fewer observations while producing physically meaningful solutions that satisfy the underlying governing equations.

Despite these developments, the application of PINNs to greenhouse climate modelling remains relatively limited. Existing studies have primarily focused on climate forecasting, single-variable prediction, or hybrid modelling strategies \cite{Choi2026,Liu2025}. Relatively little attention has been devoted to the simultaneous reconstruction of coupled greenhouse temperature and humidity together with the identification of unknown physical parameters within a unified learning framework. Furthermore, previous studies generally evaluate model performance using a single validation strategy, making it difficult to distinguish interpolation capability from true temporal generalization.

The present study addresses these limitations by developing a coupled PINN for simultaneous greenhouse temperature and humidity reconstruction from sparse observations. Unlike conventional supervised learning approaches, the proposed framework incorporates coupled physical constraints directly into the network optimization while treating the unknown physical coefficients as trainable variables. Furthermore, the proposed methodology is evaluated under two complementary experimental settings. Experiment~A investigates climate reconstruction from randomly sampled observations within the available data, whereas Experiment~B considers chronological temporal extrapolation by predicting an unseen future time interval.
Together with comparisons against a fully connected neural network, LSTM, and GRU architectures, these complementary evaluation protocols provide a more comprehensive assessment of reconstruction accuracy, predictive robustness, and model generalization under realistic greenhouse operating conditions.

%=========================================================
\section{Reduced Coupled Greenhouse Climate Model}
\label{sec:model}
%=========================================================

The objective of this study is to reconstruct the evolution of the greenhouse microclimate from sparse observations while simultaneously estimating the dominant physical parameters governing the system. Since the primary focus is inverse modelling rather than high-fidelity greenhouse simulation, a reduced-order dynamical model is adopted. The model preserves the dominant thermo-hydrological interactions within the greenhouse while remaining sufficiently simple for efficient parameter estimation through physics-informed learning.

Let $T(t)$ and $H(t)$ denote the indoor air temperature ($^\circ$C) and indoor relative humidity (\%), respectively. The greenhouse dynamics are influenced by measurable environmental forcing variables, including the outdoor temperature $T_{\mathrm{out}}(t)$, outdoor relative humidity $H_{\mathrm{out}}(t)$, solar radiation $R(t)$, and wind speed $V(t)$. These variables constitute the external inputs of the proposed model. Instead of modelling every physical process occurring inside the greenhouse, the reduced formulation captures the dominant mechanisms responsible for heat exchange, moisture transport, ventilation, and solar forcing.

Motivated by classical greenhouse energy and moisture balance models \cite{Bennis2008,VanStraten2010}, the coupled greenhouse dynamics are described by the following system of ordinary differential equations:

\begin{align}
\frac{dT}{dt}
&=
a_{1}\left(T_{\mathrm{out}}-T\right)
+a_{2}R
-a_{3}V\left(T-T_{\mathrm{out}}\right) \notag
\\
&\qquad+a_{4}\left(H-H_{\mathrm{out}}\right)
+a_{5}RV,
\label{eq:Tmodel}
\\[2mm]
\frac{dH}{dt} 
&=
b_{1}\left(H_{\mathrm{out}}-H\right)
+b_{2}R
-b_{3}V\left(H-H_{\mathrm{out}}\right) \notag
\\
&\qquad+b_{4}\left(T_{\mathrm{out}}-T\right)
+b_{5}R\left(T-T_{\mathrm{out}}\right),
\label{eq:Hmodel}
\end{align}
where the unknown coefficients
\[
\Theta=
\{
a_1,a_2,a_3,a_4,a_5,
b_1,b_2,b_3,b_4,b_5
\}
\]
represent the effective thermal and moisture exchange parameters of the reduced greenhouse system. Unlike conventional parameter calibration procedures, these coefficients are treated as trainable variables and are estimated automatically during PINN training.

The physical interpretation of the model coefficients is summarized below.

\begin{itemize}

\item $a_1$ represents the thermal relaxation between the indoor and outdoor environments.

\item $a_2$ quantifies the heating effect induced by solar radiation.

\item $a_3$ characterizes ventilation-driven heat exchange.

\item $a_4$ models the coupling between indoor humidity and temperature.

\item $a_5$ represents the interaction between solar radiation and wind-driven heat transfer.

\item $b_1$ represents the moisture exchange between indoor and outdoor air.

\item $b_2$ accounts for radiation-related moisture generation.

\item $b_3$ characterizes ventilation-induced moisture transport.

\item $b_4$ describes the influence of temperature differences on indoor humidity.

\item $b_5$ models the combined effect of solar radiation and thermal gradients on moisture dynamics.

\end{itemize}

The reduced model is intentionally designed to balance physical interpretability and computational efficiency. Although it does not explicitly represent every greenhouse process, it captures the dominant thermo-hydrological interactions required for reliable state reconstruction and parameter identification from sparse observations. This reduced formulation therefore provides an appropriate physical backbone for the proposed PINN.

It should be emphasized that the governing equations remain identical throughout this study. The distinction between the two experimental settings does not arise from different mathematical models but from different data partition strategies. Experiment~A employs randomly sampled observations distributed over the entire observation period, thereby assessing interpolation capability under sparse sensing conditions. In contrast, Experiment~B adopts a chronological train--test split, where the model is trained using the initial portion of the measurements and evaluated on a previously unseen future time interval. Consequently, Experiment~B constitutes a substantially more challenging temporal extrapolation problem while preserving the same governing physical model.

%=========================================================
\section{Physics-Informed Neural Network Formulation}
%\label{sec:pinn}
%=========================================================

The reduced greenhouse model introduced in Section~\ref{sec:model} describes the physical evolution of the indoor temperature and relative humidity through a system of coupled ordinary differential equations. In practice, however, the greenhouse states are observed only at discrete time instants and are often affected by missing measurements and sensor noise. Consequently, the governing equations and the available observations are jointly incorporated into a PINN, enabling simultaneous state reconstruction and parameter identification.

Unlike conventional supervised learning methods, the proposed framework does not rely solely on observational data. Instead, the neural network is trained to satisfy both the measured observations and the governing physical laws. The unknown physical coefficients introduced in Section~\ref{sec:model} are treated as trainable variables and are estimated together with the network parameters through a unified optimization procedure.

The neural network receives five input variables,
\[
\mathbf{x}
=
\left[
t,
T_{\mathrm{out}},
H_{\mathrm{out}},
R,
V
\right]^T,
\]
where \(t\) denotes time, \(T_{\mathrm{out}}\) and \(H_{\mathrm{out}}\) represent the outdoor temperature and humidity, \(R\) denotes the normalized solar radiation, and \(V\) represents the normalized wind speed. The network predicts the normalized greenhouse climate states
\[
\hat{\mathbf{y}}
=
\left[
\hat{T},
\hat{H}
\right]^T,
\]
corresponding to the indoor temperature and indoor relative humidity.

Accordingly, the mapping learned by the neural network can be expressed as
\[
\mathcal{N}_{\theta}:
\mathbb{R}^{5}
\rightarrow
\mathbb{R}^{2},
\]
where \(\theta\) denotes the trainable neural-network parameters. During training, both the network parameters \(\theta\) and the unknown physical coefficients
\[
\Theta=
\{
a_1,\ldots,a_5,
b_1,\ldots,b_5
\}
\]
are optimized simultaneously.

\subsection{Physics Residuals}
The governing differential equations are enforced through automatic differentiation. Let
\[
\hat{T}(t),
\qquad
\hat{H}(t)
\]
denote the network predictions in physical units. Their temporal derivatives are computed directly using automatic differentiation,
\[
\frac{d\hat{T}}{dt},
\qquad
\frac{d\hat{H}}{dt},
\]
which eliminates the need for finite-difference approximations and enables exact differentiation of the neural-network output with respect to time.

\begin{align}
\mathcal{R}_T
&=
\frac{d\hat{T}}{dt}
-
f_T
(
\hat{T},
\hat{H},
T_{\rm out},
H_{\rm out},
R,
V
), \notag
\\
\mathcal{R}_H
&=
\frac{d\hat{H}}{dt}
-
f_H
(
\hat{T},
\hat{H},
T_{\rm out},
H_{\rm out},
R,
V
),\notag
\end{align}
where $f_T$ and $f_H$ are given by Eqs.~(\ref{eq:Tmodel})--(\ref{eq:Hmodel}).

%----------------------------------------------------------
\vspace{2mm}
\textbf{Observation loss.}
%----------------------------------------------------------
The observation loss enforces agreement between the network predictions and the available greenhouse measurements. Let
$\mathcal{D}=\{(\mathbf{x}_i,\mathbf{y}_i)\}_{i=1}^{N}$
denote the set of training observations, where
$\mathbf{y}_i=(T_i,H_i)$
contains the measured indoor temperature and relative humidity. The observation loss is defined as
\begin{equation}
\mathcal{L}_{\mathrm{data}}
=
\frac{1}{N}
\sum_{i=1}^{N}
\left[
(\hat T_i-T_i)^2
+
(\hat H_i-H_i)^2
\right].
\notag%\label{eq:data}
\end{equation}
Minimizing $\mathcal{L}_{\mathrm{data}}$ ensures that the predicted greenhouse states remain consistent with the available measurements.

%----------------------------------------------------------
\vspace{2mm}
\textbf{Physics loss.}
%----------------------------------------------------------
To ensure that the reconstructed climate states satisfy the governing greenhouse dynamics, the residuals of Eqs.~(\ref{eq:Tmodel})--(\ref{eq:Hmodel}) are incorporated into the learning process. The physics loss is computed as
\begin{equation}
\mathcal{L}_{\mathrm{phys}}
=
\frac{1}{M}
\sum_{j=1}^{M}
\left[
\mathcal{R}_{T,j}^{2}
+
\mathcal{R}_{H,j}^{2}
\right],
\notag%\label{eq:phys}
\end{equation}
where $M$ denotes the number of collocation points and
$\mathcal{R}_{T}$ and $\mathcal{R}_{H}$ are the temperature and humidity residuals, respectively. Minimizing this term encourages the neural-network predictions to satisfy the governing physical laws throughout the computational domain rather than only at the observation locations.

%----------------------------------------------------------
\vspace{2mm}
\textbf{Initial condition loss.}
%----------------------------------------------------------
The initial condition is incorporated explicitly to ensure consistency between the reconstructed solution and the observed greenhouse state at the beginning of the observation period. The corresponding loss is defined as

\begin{equation}
\mathcal{L}_{\mathrm{init}}
=
(\hat T(0)-T_0)^2
+
(\hat H(0)-H_0)^2,\notag
%\label{eq:init}
\end{equation}

where $T_0$ and $H_0$ denote the measured initial indoor temperature and relative humidity, respectively.

%----------------------------------------------------------
\vspace{2mm}
\textbf{Overall training objective.}
%----------------------------------------------------------
The proposed PINN is trained by minimizing a weighted combination of the observation, physics, and initial-condition losses,
\begin{equation}
\mathcal{L}
=
\lambda_{\mathrm{data}}\mathcal{L}_{\mathrm{data}}
+
\lambda_{\mathrm{phys}}\mathcal{L}_{\mathrm{phys}}
+
\lambda_{\mathrm{init}}\mathcal{L}_{\mathrm{init}}, \notag
%\label{eq:total}
\end{equation}
where
$\lambda_{\mathrm{data}}$,
$\lambda_{\mathrm{phys}}$,
and
$\lambda_{\mathrm{init}}$
control the relative contributions of the three loss components during optimization.

In this study, the weighting coefficients are chosen as
$\lambda_{\mathrm{data}}=1$,
$\lambda_{\mathrm{phys}}=0.01$,
and
$\lambda_{\mathrm{init}}=1$,
which were found to provide a suitable balance between measurement fidelity and physical consistency. These values correspond to the final 

\section{Experimental Setup}
%\label{sec:experiment}
\subsection{Greenhouse Dataset}

The proposed framework is evaluated using the publicly available Greenhouse Shirvan Iran Dataset \cite{Salehi2024Dataset}, which contains six months of greenhouse measurements collected between November 2023 and April 2024 in Shirvan, North Khorasan Province, Iran. The dataset was acquired from an automated greenhouse monitoring system and includes indoor and outdoor temperature, indoor and outdoor relative humidity, wind speed, light intensity, and greenhouse actuator states. In this study, indoor temperature and indoor relative humidity are considered as the target variables to be reconstructed, whereas outdoor temperature, outdoor relative humidity, solar radiation (light intensity), and wind speed are employed as external forcing variables in the reduced greenhouse model.

To facilitate stable neural-network training, each input variable is normalized using statistics computed exclusively from the training dataset. The same normalization parameters are subsequently applied to the testing dataset, thereby preventing information leakage between the training and testing phases.

\begin{figure*}[!t]
\centering
\includegraphics[width=0.95\textwidth]{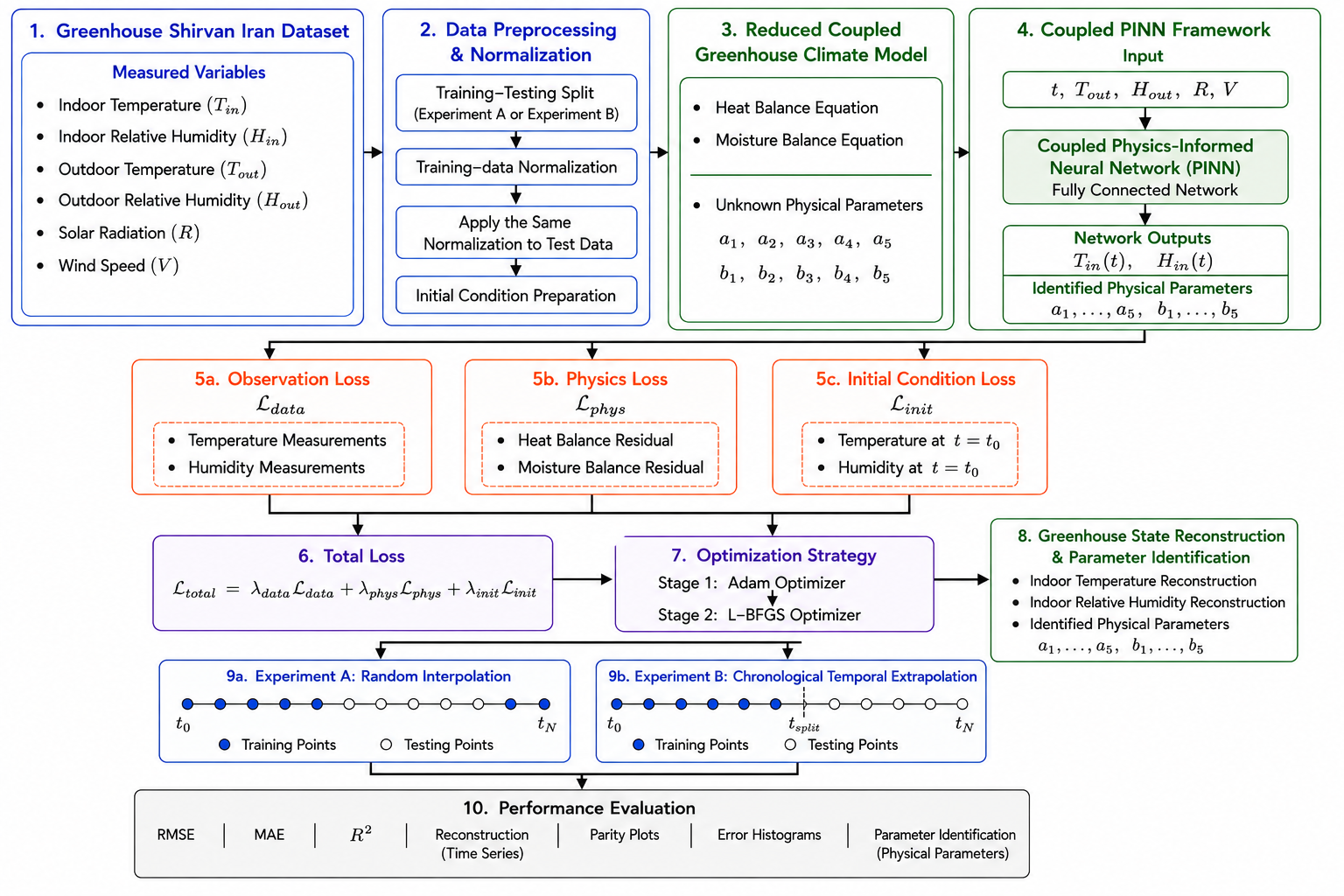}
\caption{Overall workflow of the proposed coupled PINN framework, including data preprocessing, physics-informed learning, greenhouse climate reconstruction, physical parameter identification, and experimental evaluation.}
\label{fig:framework}
\end{figure*}

Fig.~\ref{fig:framework} illustrates the overall workflow of the proposed coupled PINN framework. The greenhouse observations are first preprocessed and normalized before being supplied to the coupled PINN together with the reduced greenhouse climate model. During training, the network simultaneously reconstructs indoor temperature and relative humidity while identifying the unknown physical parameters by jointly minimizing the observation, physics, and initial-condition losses using a two-stage Adam--L-BFGS optimization strategy. After training, the learned model provides physically consistent greenhouse climate reconstruction together with the identified physical parameters, which are subsequently evaluated under both random interpolation (Experiment~A) and chronological temporal extrapolation (Experiment~B).

\subsection{Network Architecture}

The proposed PINN employs a fully connected multilayer perceptron consisting of four hidden layers with 64 neurons per layer. Hyperbolic tangent (Tanh) activation functions are used throughout the hidden layers because of their smooth differentiability, which is advantageous for automatic differentiation.

The network receives five input variables,
\[
(t,
T_{\rm out},
H_{\rm out},
R,
V),
\]
and simultaneously predicts the normalized indoor temperature and indoor relative humidity.
\subsection{Training Procedure}

The unknown neural-network parameters and the physical coefficients are optimized simultaneously using a two-stage optimization strategy.

An initial optimization is performed using the Adam optimizer with a learning rate of \(10^{-3}\) for 12\,000 epochs. Subsequently, the L-BFGS optimizer is employed to further minimize the composite loss and improve satisfaction of the governing physical equations.

The weighting coefficients used throughout all experiments are
\[
\lambda_{\rm data}=1,\qquad
\lambda_{\rm phys}=0.01,\qquad
\lambda_{\rm init}=1.
\]
These values were found to provide an appropriate balance between observational accuracy and physical consistency.

\subsection{Experimental Protocols}

To comprehensively evaluate the proposed framework, two complementary experimental protocols are considered. The first assesses interpolation performance under randomly sampled observations, whereas the second evaluates predictive generalization through chronological temporal extrapolation.

\textbf{Experiment A (Random Interpolation).}
The complete dataset is randomly divided into training (70\%) and testing (30\%) subsets while preserving the overall temporal coverage of the observations. Consequently, both subsets are drawn from the same operating regime, and the experiment evaluates the ability of the proposed framework to reconstruct greenhouse climate states from sparse observations.

\textbf{Experiment B (Chronological Temporal Extrapolation).}
The dataset is partitioned chronologically, with the first 70\% of the observations used for training and the remaining 30\% reserved for testing. Since the testing interval occurs entirely after the training period, this experiment evaluates the capability of the proposed framework to predict greenhouse climate under previously unseen future operating conditions.

For both experiments, the proposed coupled PINN is compared with three representative data-driven models: a fully connected neural network (Pure NN), a LSTM network, and a GRU network. All models are trained and evaluated using the corresponding data partitions under identical implementation settings to ensure a fair comparison.

%=========================================================
\section{Results and Discussion}
%\label{sec:results}
%=========================================================

\subsection{Experiment A: Random Interpolation}

\subsubsection*{Training Convergence}

\begin{figure}%[!t]
\centering
\subfloat[\label{fig:PINN_training_loss}]{\includegraphics[width=0.48\linewidth]{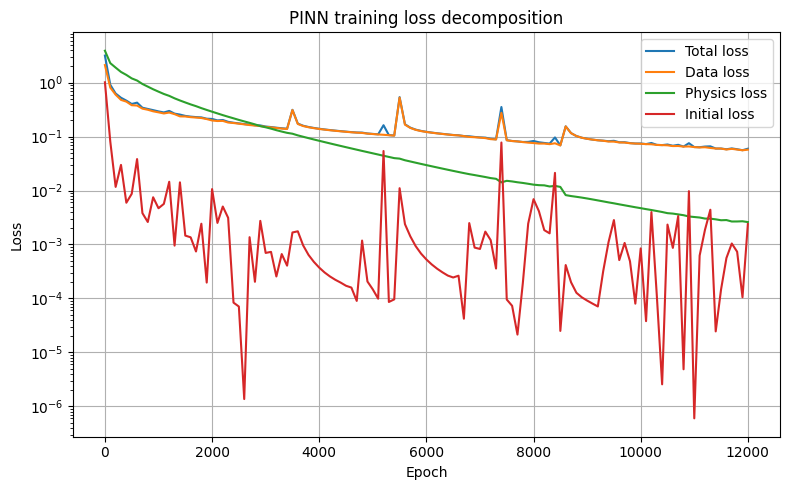}}
\subfloat[\label{fig:Training_loss_convergence}]{\includegraphics[width=0.48\linewidth]{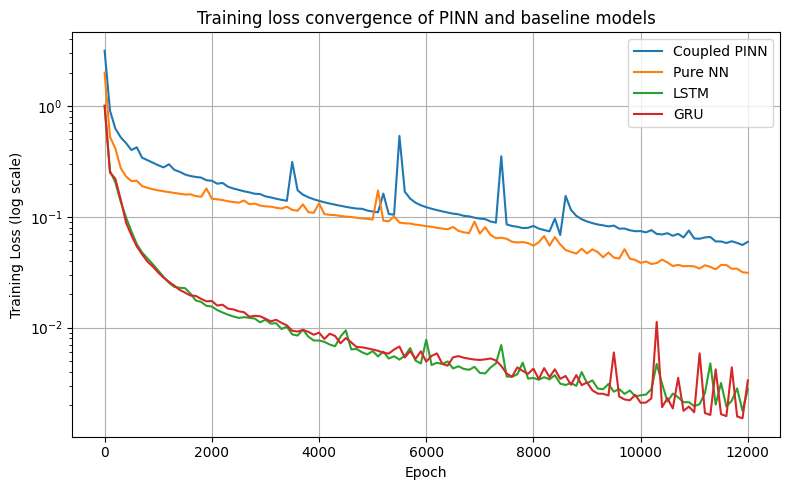}}
\caption{Training convergence for Experiment~A.
(a) Evolution of the total, observation, physics, and initial-condition losses of the proposed coupled PINN.
(b) Comparison of the optimization histories of the coupled PINN, Pure NN, LSTM, and GRU models, illustrating their convergence behaviour during training.}
\label{fig:ExpA_loss}
\end{figure}

The convergence histories of the proposed coupled PINN and the competing
data-driven models are presented in Figs.~\ref{fig:ExpA_loss}(a)--(b).
Fig.~\ref{fig:ExpA_loss}(a) illustrates the evolution of the individual
loss components of the coupled PINN during optimization. The observation
loss decreases rapidly during the early training stage, while the
physics loss is progressively reduced as the network learns solutions
that simultaneously satisfy the measurement data and the governing
greenhouse equations. The initial-condition loss rapidly approaches
zero, demonstrating consistency with the measured initial greenhouse
state. After 12\,000 Adam iterations, an L-BFGS refinement stage further
reduces the total loss to approximately $4.05\times10^{-2}$, indicating
stable convergence of the coupled optimization problem.

Fig.~\ref{fig:ExpA_loss}(b) compares the optimization histories of the
coupled PINN, Pure NN, LSTM, and GRU models. The Pure NN converges most
rapidly, requiring only 70.29~s of training. Owing to the simultaneous
optimization of the neural-network weights and unknown physical
coefficients, the coupled PINN requires 405.39~s. The recurrent
architectures are computationally more demanding, with training times
of 1005.70~s and 624.40~s for the LSTM and GRU models, respectively.
Although the PINN requires longer optimization than the Pure NN, it
provides physically interpretable parameters while maintaining stable
training behaviour.

\subsubsection*{Greenhouse Climate Reconstruction}

\begin{figure}%[!t]
\centering
\subfloat[\label{fig:Iranian_greenhouse_temperature}]{\includegraphics[width=0.48\linewidth]{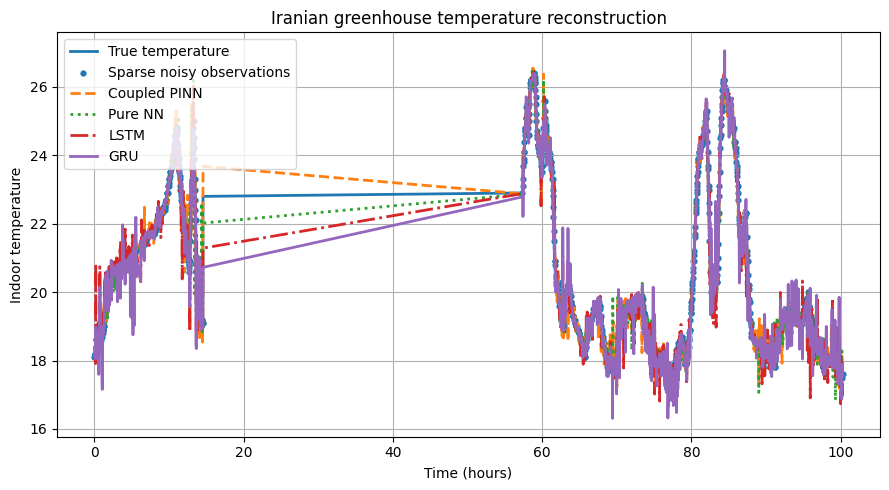}}
\subfloat[\label{fig:Iranian_greenhouse_humidity}]{\includegraphics[width=0.48\linewidth]{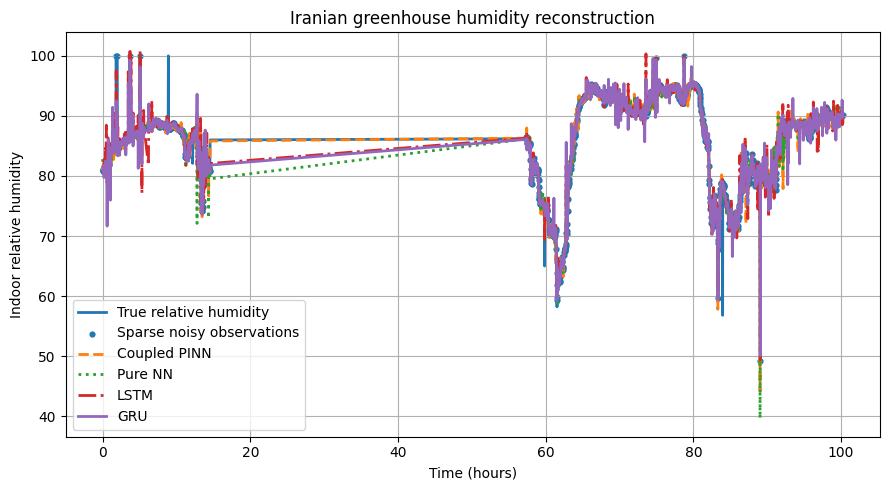}}
\caption{Greenhouse climate reconstruction in Experiment~A.
(a) Reconstruction of indoor temperature.
(b) Reconstruction of indoor relative humidity.}
\label{fig:ExpA_reconstruction}
\end{figure}

Fig.~\ref{fig:ExpA_reconstruction}(a)--(b) compare the reconstructed
indoor temperature and relative humidity obtained using the four
competing models. Overall, all approaches successfully recover the
dominant temporal evolution of the greenhouse climate under randomly
sampled observations. The coupled PINN closely follows the measured
temperature profile throughout the observation interval, accurately
capturing both heating and cooling phases. Humidity reconstruction is
more challenging because of stronger short-term fluctuations; however,
all models preserve the principal trends, with the PINN and recurrent
networks providing consistent reconstructions over most of the testing
interval.

\subsubsection*{Quantitative Performance Evaluation}

The quantitative prediction performance of the four competing models is
summarized in Table~\ref{tab:ExpA_metrics}. For indoor temperature, the
coupled PINN achieves the lowest RMSE (0.4495$^\circ$C) and the highest
coefficient of determination ($R^2=0.9636$), demonstrating the most
accurate temperature reconstruction. The Pure NN produces comparable
performance, whereas the recurrent models exhibit slightly larger
prediction errors. For indoor humidity, the LSTM achieves the best
prediction accuracy with an RMSE of 2.3563\% and an $R^2$ value of
0.9063. Nevertheless, the coupled PINN remains highly competitive,
achieving an RMSE of 2.4818\% and an $R^2$ value of 0.8960 while
simultaneously identifying the unknown physical parameters of the
reduced greenhouse model. These results indicate that incorporating
physical constraints substantially improves temperature reconstruction
without sacrificing humidity prediction accuracy.

\begin{table}[!t]
\caption{Quantitative performance comparison of the coupled PINN and competing data-driven models for Experiment~A.}
\label{tab:ExpA_metrics}
\scriptsize
\centering
\renewcommand{\arraystretch}{1.15}
\begin{tabular}{lcccc}
\hline
Method & Time (s) & Metric & Temperature ($^\circ\mathrm{C}$) & Humidity ($\%$) \\
\hline
\multirow{3}{*}{Pure NN}
& \multirow{3}{*}{70.29}
& RMSE & 0.4716 & 2.6432 \\
& & MAE & 0.2955 & 1.1287 \\
& & $R^2$ & 0.9599 & 0.8821 \\
\hline

\multirow{3}{*}{Coupled PINN}
& \multirow{3}{*}{405.39}
& RMSE & \textbf{0.4495} & 2.4818 \\
& & MAE & 0.3045 & 1.1748 \\
& & $R^2$ & \textbf{0.9636} & 0.8960 \\
\hline

\multirow{3}{*}{LSTM}
& \multirow{3}{*}{1005.70}
& RMSE & 0.5793 & \textbf{2.3563} \\
& & MAE & 0.3839 & 1.3668 \\
& & $R^2$ & 0.9395 & \textbf{0.9063} \\
\hline

\multirow{3}{*}{GRU}
& \multirow{3}{*}{624.40}
& RMSE & 0.6447 & 2.5272 \\
& & MAE & 0.4392 & 1.4944 \\
& & $R^2$ & 0.9250 & 0.8922 \\
\hline
\end{tabular}
\end{table}

\subsubsection*{Prediction Accuracy and Error Analysis}

\begin{figure}[!t]
\centering

\subfloat[\label{fig:temp_parity}]{
\includegraphics[width=0.48\linewidth]{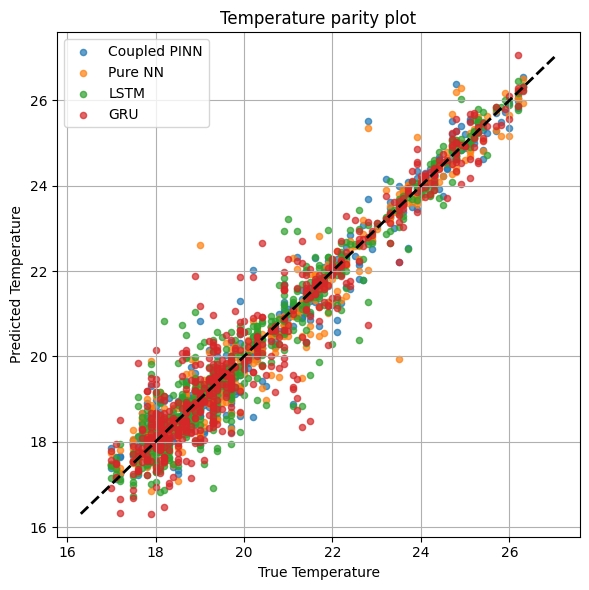}}
\hfill
\subfloat[\label{fig:hum_parity}]{
\includegraphics[width=0.48\linewidth]{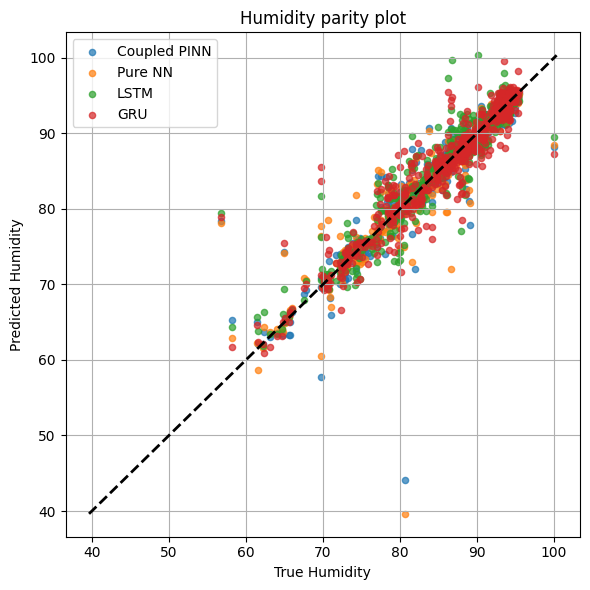}}

%\vspace{1mm}

\subfloat[\label{fig:temp_error}]{
\includegraphics[width=0.48\linewidth]{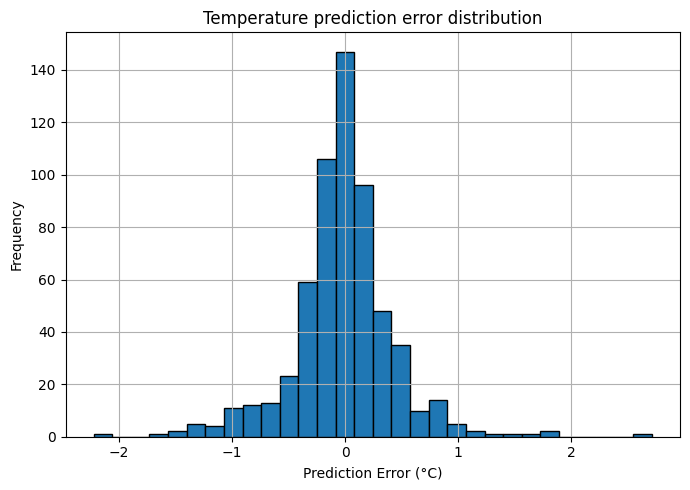}}
\hfill
\subfloat[\label{fig:hum_error}]{
\includegraphics[width=0.48\linewidth]{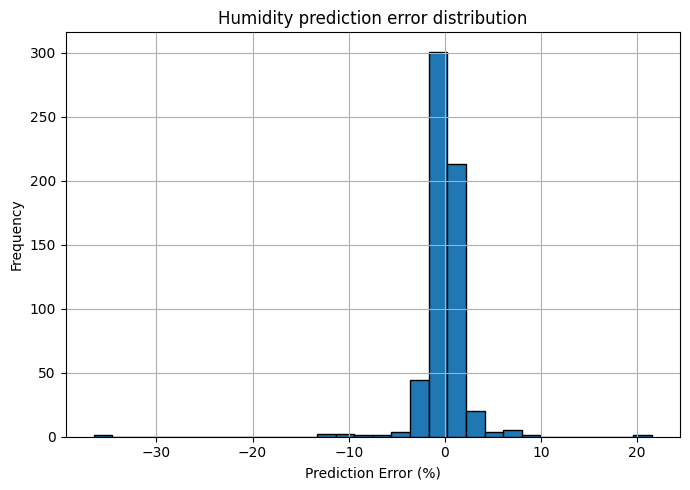}}

\caption{Prediction accuracy analysis for Experiment~A.
(a) Temperature parity plot.
(b) Humidity parity plot.
(c) Temperature prediction error distribution.
(d) Humidity prediction error distribution.}
\label{fig:ExpA_accuracy}
\end{figure}

Fig.~\ref{fig:ExpA_accuracy}(a)--(d) further evaluate the prediction
quality using parity plots and prediction-error histograms. The parity
plots reveal that the predicted temperature and humidity values are
closely clustered around the 45$^\circ$ reference line, confirming good
agreement between the reconstructed and measured greenhouse states.
Consistent with Table~\ref{tab:ExpA_metrics}, temperature predictions
exhibit slightly smaller dispersion than humidity predictions. The error
histograms are approximately centred around zero, indicating that the
proposed framework does not introduce significant systematic prediction
bias.

\subsubsection*{Estimated Physical Parameters}

\begin{figure}%[!t]
\centering
\includegraphics[width=0.65\linewidth]{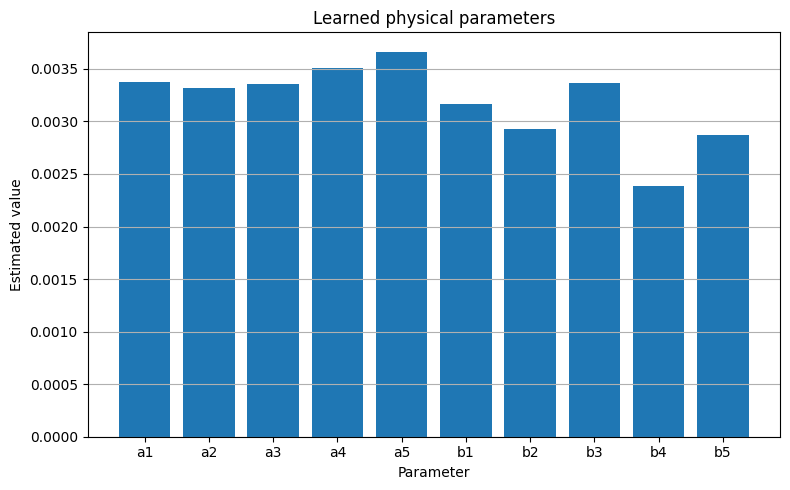}
\caption{Estimated thermal and moisture exchange parameters identified by the coupled PINN for Experiment~A.}
\label{fig:ExpA_parameters}
\end{figure}

The physical coefficients identified by the coupled PINN are presented
in Fig.~\ref{fig:ExpA_parameters}. All estimated parameters remain
positive and are of similar magnitude, indicating stable parameter
identification throughout the optimization process. The estimated values
characterize the dominant heat and moisture exchange mechanisms
represented by the reduced greenhouse model. Unlike purely data-driven
methods, the proposed PINN simultaneously reconstructs greenhouse
climate states and estimates physically interpretable model parameters,
thereby improving the transparency of the learned model.

%=========================================================
\subsection{Experiment B: Chronological Temporal Extrapolation}
%=========================================================

\subsubsection*{Training Convergence}

\begin{figure}
\centering
\subfloat[\label{fig:PINN_training_lossB}]{\includegraphics[width=0.48\linewidth]{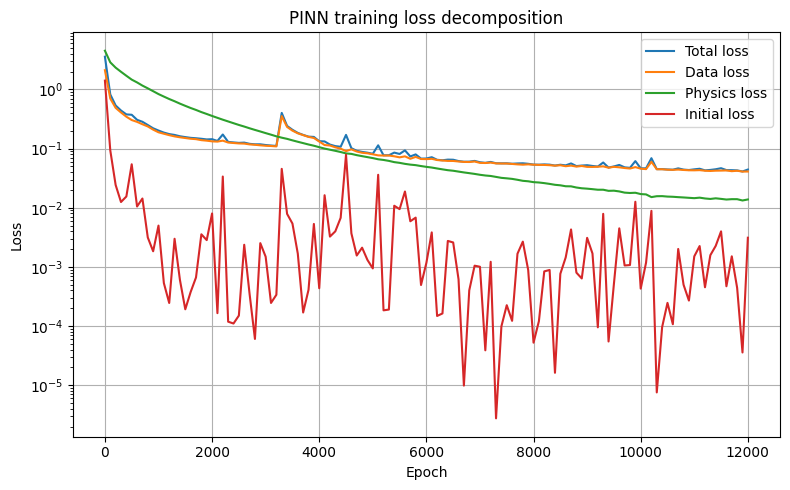}}
\subfloat[\label{fig:Training_loss_convergenceB}]{\includegraphics[width=0.48\linewidth]{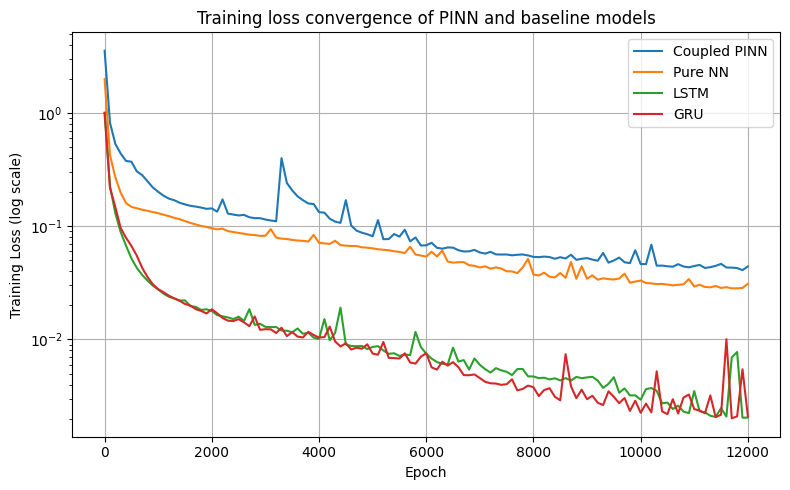}}
\caption{Training convergence for Experiment~B.
(a) Evolution of the total, observation, physics, and initial-condition losses of the proposed coupled PINN.
(b) Comparison of the optimization histories of the coupled PINN, Pure NN, LSTM, and GRU models, illustrating their convergence behaviour during training.}
\label{fig:ExpB_loss}
\end{figure}

The convergence behaviour of the coupled PINN and the competing
data-driven models under the chronological temporal extrapolation
setting is presented in Fig.~\ref{fig:ExpB_loss}(a)--(b). Similar to
Experiment~A, the observation, physics, and initial-condition losses
decrease steadily throughout the optimization process, indicating that
the proposed PINN successfully balances observational accuracy and
physical consistency. After 12\,000 Adam iterations, the subsequent
L-BFGS refinement further improves the optimization, yielding a final
total loss of approximately $3.45\times10^{-2}$.

Fig.~\ref{fig:ExpB_loss}(b) compares the optimization histories of the
four competing models. The Pure NN again requires the shortest training
time (73.51~s), whereas the coupled PINN converges in 333.07~s owing to
the simultaneous optimization of the neural-network parameters and the
unknown physical coefficients. The sequential architectures require
considerably longer computational times, with the LSTM and GRU models
requiring 1218.61~s and 650.57~s, respectively. Despite the increased
optimization cost, the coupled PINN exhibits stable convergence
throughout the training process.

\subsubsection*{Future Greenhouse Climate Prediction}

\begin{figure}
\centering
\subfloat[\label{fig:Iranian_greenhouse_temperatureB}]{\includegraphics[width=0.48\linewidth]{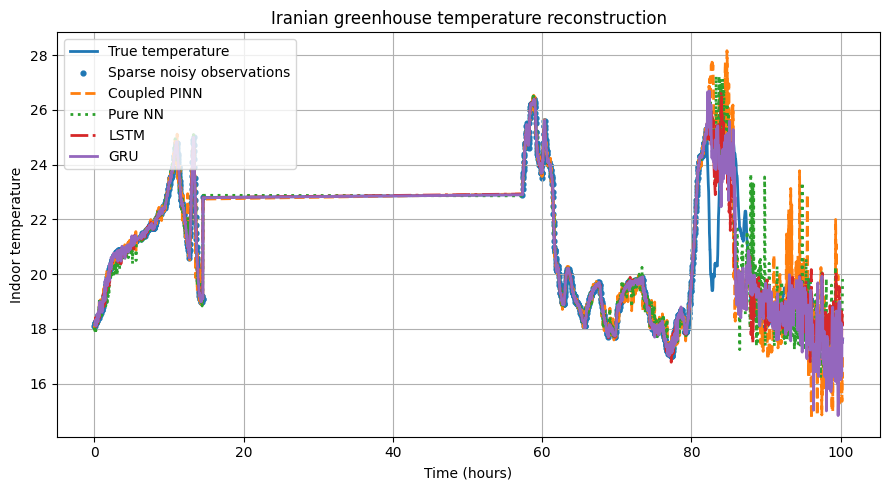}}
\subfloat[\label{fig:Iranian_greenhouse_humidityB}]{\includegraphics[width=0.48\linewidth]{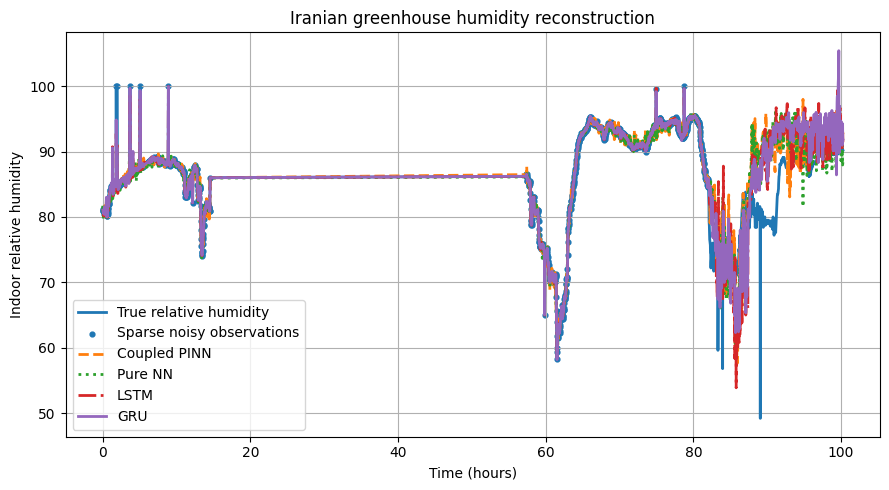}}
\caption{Greenhouse climate prediction in Experiment~B.
(a) Indoor temperature prediction.
(b) Indoor relative humidity prediction.}
\label{fig:ExpB_reconstruction}
\end{figure}

The prediction results over the unseen future testing interval are
presented in Figs.~\ref{fig:ExpB_reconstruction}(a)--(b). Compared with
Experiment~A, the prediction task is considerably more challenging
because the testing observations occur entirely after the training
period. Consequently, all competing models experience noticeable
performance degradation.

Fig.~\ref{fig:ExpB_reconstruction}(a) shows the predicted indoor
temperature. Although the dominant temporal trend is captured by all
models, larger deviations are observed during rapid temperature
variations. Fig.~\ref{fig:ExpB_reconstruction}(b) presents the
corresponding indoor humidity predictions. Relative humidity exhibits
substantially stronger temporal variability, making long-term prediction
more difficult than temperature reconstruction. Nevertheless, all four
models preserve the principal evolution of the greenhouse climate over
the forecasting horizon.

\subsubsection*{Quantitative Performance Evaluation}

The quantitative prediction performance under chronological temporal
extrapolation is summarized in Table~\ref{tab:ExpB_metrics}. As
anticipated, all models exhibit larger prediction errors than in
Experiment~A because the testing interval lies completely outside the
training period.

For indoor temperature prediction, the LSTM achieves the lowest RMSE
(1.8476$^\circ$C) and the highest coefficient of determination
($R^2=0.4383$), followed closely by the GRU model. The coupled PINN and
Pure NN produce comparable temperature predictions but experience a
larger degradation under temporal extrapolation.

Humidity prediction proves even more challenging. Although all models
show reduced performance, the GRU achieves the smallest RMSE
(6.7354\%) and is the only model producing a positive coefficient of
determination ($R^2=0.0137$). The remaining models exhibit negative
$R^2$ values, indicating the increased difficulty of forecasting indoor
humidity over an unseen future period.

Overall, the results demonstrate that chronological extrapolation is
considerably more demanding than random interpolation for every
competing model. Nevertheless, the proposed coupled PINN remains capable
of producing physically consistent greenhouse climate predictions while
simultaneously identifying the unknown physical parameters of the
reduced greenhouse model.

\begin{table}[!t]
\caption{Quantitative performance comparison of the coupled PINN and competing data-driven models for Experiment~B.}
\label{tab:ExpB_metrics}
\scriptsize
\centering
\renewcommand{\arraystretch}{1.15}
\begin{tabular}{lcccc}
\hline
Method & Time (s) & Metric & Temperature ($^\circ\mathrm{C}$) & Humidity ($\%$) \\
\hline

\multirow{3}{*}{Pure NN}
& \multirow{3}{*}{73.51}
& RMSE & 2.0450 & 7.1357 \\
& & MAE  & 1.5029 & 5.6675 \\
& & $R^2$ & 0.3119 & -0.1071 \\
\hline

\multirow{3}{*}{Coupled PINN}
& \multirow{3}{*}{333.07}
& RMSE & 2.2300 & 7.3725 \\
& & MAE  & 1.4824 & 5.9149 \\
& & $R^2$ & 0.1817 & -0.1817 \\
\hline

\multirow{3}{*}{LSTM}
& \multirow{3}{*}{1218.61}
& RMSE & \textbf{1.8476} & 7.1768 \\
& & MAE  & \textbf{1.2778} & 5.7670 \\
& & $R^2$ & \textbf{0.4383} & -0.1199 \\
\hline

\multirow{3}{*}{GRU}
& \multirow{3}{*}{650.57}
& RMSE & 1.9108 & \textbf{6.7354} \\
& & MAE  & 1.4087 & \textbf{5.6594} \\
& & $R^2$ & 0.3992 & \textbf{0.0137} \\
\hline
\end{tabular}
\end{table}

\subsubsection*{Prediction Accuracy and Error Analysis}

\begin{figure}[!t]
\centering

\subfloat[\label{fig:temp_parityB}]{
\includegraphics[width=0.48\linewidth]{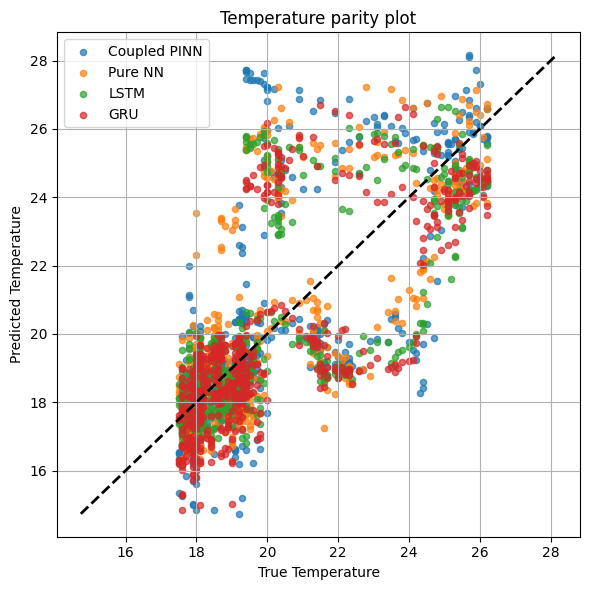}}
\hfill
\subfloat[\label{fig:hum_parityB}]{
\includegraphics[width=0.48\linewidth]{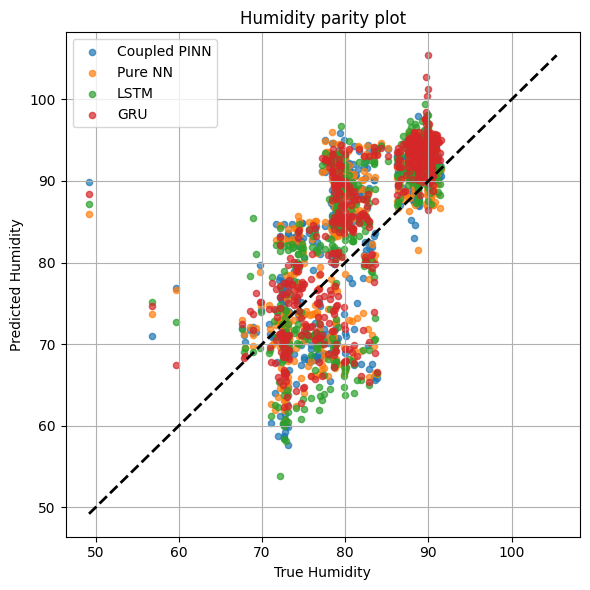}}

%\vspace{1mm}

\subfloat[\label{fig:temp_errorB}]{
\includegraphics[width=0.48\linewidth]{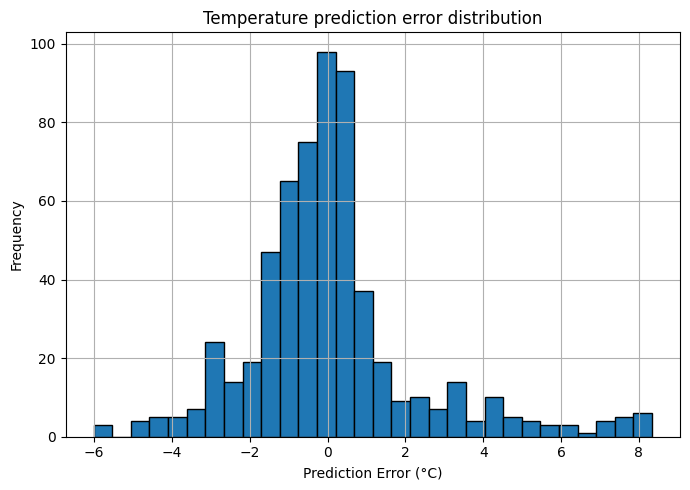}}
\hfill
\subfloat[\label{fig:hum_errorB}]{
\includegraphics[width=0.48\linewidth]{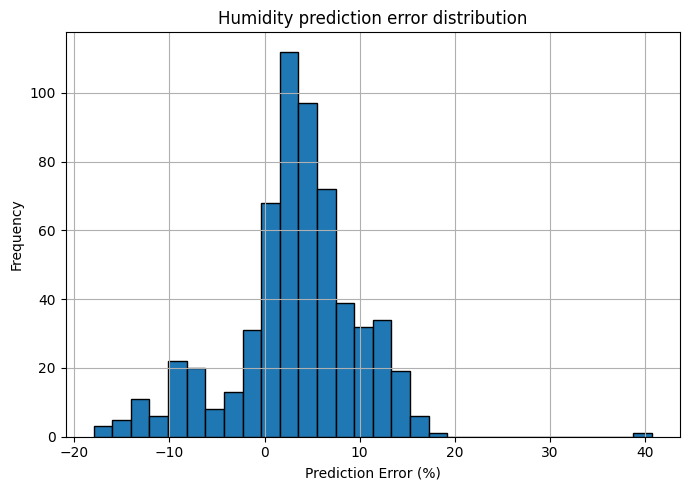}}

\caption{Prediction accuracy analysis for Experiment~B.
(a) Temperature parity plot.
(b) Humidity parity plot.
(c) Temperature prediction error distribution.
(d) Humidity prediction error distribution.}
\label{fig:ExpB_accuracy}
\end{figure}

Fig.~\ref{fig:ExpB_accuracy}(a)--(d) provide further insight into the
prediction accuracy of the four competing models. Compared with
Experiment~A, the parity plots exhibit noticeably larger scatter around
the 45$^\circ$ reference line, reflecting the increased uncertainty
associated with chronological temporal extrapolation. This behaviour is
particularly evident for humidity prediction, where abrupt fluctuations
occur throughout the testing interval.

The prediction-error distributions shown in
Figs.~\ref{fig:ExpB_accuracy}(c)--(d) remain centred approximately
around zero but display considerably larger variance than those
observed in Experiment~A. These wider error distributions are
consistent with the larger RMSE values reported in
Table~\ref{tab:ExpB_metrics} and illustrate the increased difficulty of
forecasting greenhouse climate beyond the temporal range represented
during training.

\subsubsection*{Estimated Physical Parameters}

\begin{figure}
\centering
\includegraphics[width=0.65\linewidth]{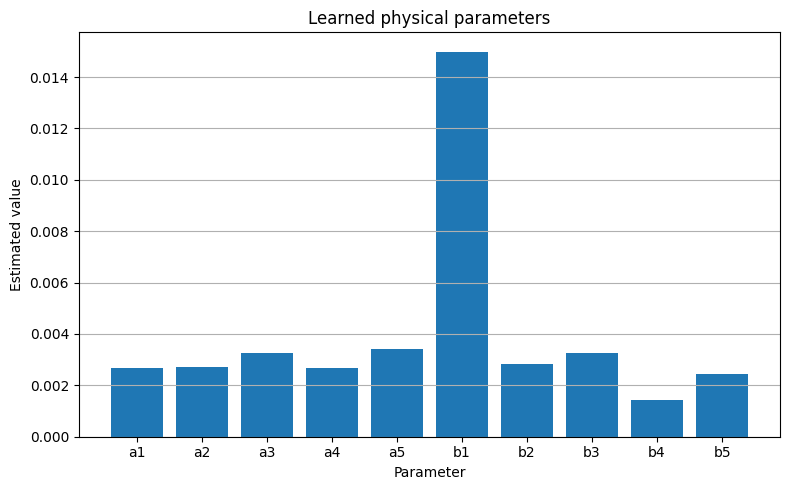}
\caption{Estimated thermal and moisture exchange parameters identified by the coupled PINN for Experiment~B.}
\label{fig:ExpB_parameters}
\end{figure}

The physical coefficients identified by the coupled PINN for
Experiment~B are presented in Fig.~\ref{fig:ExpB_parameters}. Although
the estimated parameter values differ from those obtained in
Experiment~A, all coefficients remain physically meaningful and
numerically stable throughout the optimization process. The differences
between the two experiments reflect the distinct information available
during training rather than changes in the underlying greenhouse model.
These results demonstrate that the proposed framework is capable of
simultaneously estimating greenhouse states and effective physical
parameters under both interpolation and chronological extrapolation
settings.

%=========================================================
\subsection{Comparison of Experimental Protocols}
%=========================================================
This paper presented a coupled PINN framework for greenhouse climate state estimation using a reduced greenhouse dynamic model. The proposed approach integrates physical knowledge with observational data by embedding the governing heat- and moisture-balance equations into the neural-network training process. Unlike purely data-driven models, the proposed framework simultaneously reconstructs indoor temperature and relative humidity while identifying the unknown physical parameters governing the greenhouse dynamics, thereby providing a physically interpretable solution for intelligent greenhouse monitoring under sparse sensor measurements.

A comparison of the quantitative results obtained in the two experiments reveals a clear performance degradation for all competing models under chronological temporal extrapolation. In Experiment~A, the proposed coupled PINN achieves the best temperature reconstruction with an RMSE of 0.4495$^\circ$C and an $R^2$ value of 0.9636, while the LSTM provides the highest humidity prediction accuracy with an RMSE of 2.3563\% and an $R^2$ value of 0.9063. In contrast, the prediction errors increase substantially in Experiment~B. The best temperature performance is achieved by the LSTM (RMSE = 1.8476$^\circ$C, $R^2=0.4383$), whereas the GRU provides the most accurate humidity prediction (RMSE = 6.7354\%, $R^2=0.0137$). These results demonstrate that forecasting greenhouse climate over an unseen future interval is considerably more difficult than reconstructing randomly sampled observations from the same operating regime.

The reconstruction figures further support these quantitative observations. In Experiment~A, all models closely follow the measured temperature and humidity trajectories because the training and testing samples originate from the same temporal distribution. In Experiment~B, however, noticeable deviations appear in both temperature and humidity predictions, particularly during periods of rapid environmental variation. The parity plots exhibit larger scatter around the ideal prediction line, and the prediction-error histograms become considerably wider, indicating increased uncertainty during long-term prediction.

An important observation is that the coupled PINN maintains physically consistent predictions under both experimental settings while simultaneously estimating the unknown coefficients of the reduced greenhouse model. Although the proposed PINN does not achieve the lowest prediction error for every metric in Experiment~B, it provides an interpretable modelling framework in which greenhouse state reconstruction and physical parameter identification are performed simultaneously. In contrast, the purely data-driven models optimize predictive accuracy alone and do not provide estimates of the governing physical parameters.

Overall, the two complementary validation protocols demonstrate that interpolation and chronological extrapolation evaluate fundamentally different aspects of model performance. Experiment~A measures the capability of reconstructing greenhouse states from sparse observations distributed throughout the available data, whereas Experiment~B assesses predictive generalization under previously unseen operating conditions. Considering both protocols therefore provides a more comprehensive and realistic evaluation of greenhouse climate reconstruction than either experiment alone.

%=========================================================
\section{Conclusions}
%\label{sec:conclusion}
%=========================================================

This paper presented a coupled PINN
framework for greenhouse climate reconstruction using a reduced
greenhouse dynamic model. The proposed approach integrates physical
knowledge with observational data by embedding the governing heat and
moisture balance equations into the neural-network training process.
Unlike purely data-driven models, the proposed framework simultaneously
reconstructs indoor temperature and relative humidity while identifying
the unknown physical parameters governing the greenhouse dynamics.

The proposed methodology was evaluated using measurements collected from
a commercial greenhouse under two complementary experimental protocols.
Experiment~A considered random interpolation, where the training and
testing samples were randomly distributed throughout the observation
period. Under this setting, the coupled PINN achieved the highest
temperature reconstruction accuracy while maintaining competitive
humidity prediction performance. Experiment~B investigated chronological
temporal extrapolation by predicting greenhouse climate over a future
time interval that was completely excluded from the training process.
As expected, the forecasting task proved substantially more challenging,
and all competing models experienced noticeable performance degradation.
Nevertheless, the proposed PINN produced physically consistent climate
reconstructions and simultaneously estimated meaningful physical
coefficients under both experimental settings.

The comparative study demonstrates that interpolation and temporal
extrapolation evaluate different aspects of model performance and should
both be considered when assessing greenhouse climate reconstruction
methods. While purely data-driven models may achieve slightly lower
prediction errors for specific variables under certain conditions, the
proposed coupled PINN provides the additional advantage of maintaining
consistency with the governing physical equations and yielding
interpretable physical parameters. These characteristics improve the transparency, physical interpretability, and scientific reliability of the learned model, making the proposed framework particularly suitable for intelligent greenhouse monitoring and physics-informed environmental state estimation.

While this study considers a single real greenhouse dataset from Iran, the proposed physics-informed learning framework is inherently general and can be readily extended to other greenhouse environments. Publicly available datasets, such as the South African greenhouse dataset reported by  \cite{Hull2023}, provide valuable opportunities for future validation across diverse climatic conditions, greenhouse structures, crop types, and sensing configurations. Such cross-dataset investigations would further assess the robustness, scalability, and generalizability of the proposed framework.

Future work will focus on extending the proposed framework to more comprehensive greenhouse models by incorporating additional environmental variables, including carbon dioxide concentration, soil moisture, and soil temperature. Future studies will also investigate validation across multiple greenhouse environments, crop types, and climatic conditions together with the integration of uncertainty quantification and online adaptive learning for real-time greenhouse monitoring and intelligent climate control. Overall, the proposed framework provides a promising foundation for intelligent greenhouse monitoring, virtual sensing, digital twins \cite{Wang2023DT}, and next-generation automated greenhouse climate management, thereby supporting the development of trustworthy physics-informed artificial intelligence for precision agriculture.

%\section*{Acknowledgments}

%%%%%%%%%%%%%%%%%%%%%%%%%%%%%%%%%%% 
%\begin{IEEEbiography}[{\includegraphics[width=1in,height=1.25in,clip,keepaspectratio]{sani.jpeg}}]{Sani Biswas}
% Sani Biswas received the Ph.D. degree in Mathematics from the Indian Institute of Technology (IIT) Roorkee, India. He received the INSPIRE Scholarship from the Department of Science and Technology (DST), India.
% He is a postdoctoral fellow with the Centro de Modelamiento Matem\'atico (CMM), University of Chile, under the supervision of Prof. Joaquín Fontbona. He  also visited the Stat-Math Unit, Indian Statistical Institute (ISI), Bangalore, as a scientist.  
% His research interests include McKean--Vlasov stochastic differential equations, interacting particle systems, propagation of chaos, and stochastic numerical analysis. More recently, his research has expanded to physics-guided neural methods for stochastic dynamical systems and machine-learning-based approaches for gas sensing and wireless communication applications. He has authored and co-authored research articles in international journals.
%\end{IEEEbiography}
%\vspace{-3em}

%%%%%%%%%%%%%%%%%%%%%%%%%%%%%%%%%%% Kamal %%%%%%%%%%%%%%%%%%%%%%%%%


\begin{thebibliography}{1}
\bibliographystyle{IEEEtran}
%antiguo 1
\bibitem{VanStraten2010}
G. van Straten, G. van Willigenburg, E. van Henten, and R. van Ooteghem,
\emph{Optimal Control of Greenhouse Cultivation}.
%Boca Raton, FL, USA: 
CRC Press, 2010.



\bibitem{Shamshiri2018}
R. R. Shamshiri, J. W. Jones, K. R. Thorp, D. Ahmad, H. Che Man, and S. Taheri,
``Review of optimum temperature, humidity, and vapour pressure deficit for microclimate evaluation and control in greenhouse cultivation of tomato: A review,''
\emph{International Agrophysics},
vol.~32, no.~2,
pp.~287--302,
2018.

\bibitem{Chen2025}
S. Chen, A. Liu, F. Tang, P. Hou, Y. Lu, and P. Yuan,
``A review of environmental control strategies and models for modern agricultural greenhouses,''
\emph{Sensors},
vol.~25, no.~5,
Art.~no.~1388,
2025.

\bibitem{Mowla2023}
M. N. Mowla, N. Mowla, A. F. M. Shah, K. M. Rabie, and T. Shongwe,
``Internet of Things and Wireless Sensor Networks for Smart Agriculture Applications: A Survey,''
\emph{IEEE Access},
vol.~11,
pp.~145813--145852,
2023.

\bibitem{Friha2021}
O. Friha, M. A. Ferrag, L. Shu, L. Maglaras, and X. Wang,
``Internet of Things for the Future of Smart Agriculture:
A Comprehensive Survey of Emerging Technologies,''
\emph{IEEE/CAA Journal of Automatica Sinica},
vol.~8,
no.~4,
pp.~718--752,
2021.

\bibitem{Bennis2008}
N. Bennis, J. Duplaix, G. En\'ea, M. Haloua, and H. Youlal,
``Greenhouse climate modelling and robust control,''
\emph{Computers and Electronics in Agriculture},
vol.~61,  no.~2,
pp.~96--107,
2008.

\bibitem{Singh2017}
V. K. Singh and K. N. Tiwari,
``Prediction of greenhouse micro-climate using artificial neural network,''
\emph{Applied Ecology and Environmental Research},
vol.~15, no.~1,
pp.~767--778,
2017.

\bibitem{Yang2023}
Y. Yang, P. Gao, Z. Sun, H. Wang, M. Lu, Y. Liu, and J. Hu,
``Multistep ahead prediction of temperature and humidity in solar greenhouse based on FAM-LSTM,''
\emph{Computers and Electronics in Agriculture},
vol.~213,
Art.~no.~108261,
2023.

\bibitem{Yu2025}
J. Yu, C. Sun, J. Zhao, L. Ma, W. Zheng, Q. Xie, and X. Wei,
``Prediction and control of greenhouse temperature: Methods, applications, and future directions,''
\emph{Computers and Electronics in Agriculture},
vol.~237,
Art.~no.~110603,
2025.

\bibitem{Li2026}
B. Li, Z. Wang, Y. Guo, X. Zhou, B. Qiao, and L. Han,
``LSTM-GRU hybrid model for multi-layer microclimate prediction in solar greenhouses,''
\emph{Scientific Reports},
2026.

\bibitem{Raissi2019}
M. Raissi, P. Perdikaris, and G. E. Karniadakis,
``Physics-informed neural networks: A deep learning framework for solving forward and inverse problems involving nonlinear partial differential equations,''
\emph{Journal of Computational Physics},
vol.~378,
pp.~686--707,
2019.
\bibitem{Karniadakis2021}
G. E. Karniadakis, I. G. Kevrekidis, L. Lu, P. Perdikaris, S. Wang, and L. Yang,
``Physics-informed machine learning,''
\emph{Nature Reviews Physics},
vol.~3, no.~6,
pp.~422--440,
2021.

\bibitem{Cuomo2022}
S. Cuomo, V. S. Di Cola, F. Giampaolo, G. Rozza, M. Raissi, and F. Piccialli,
``Scientific machine learning through physics-informed neural networks: Where we are and what is next,''
\emph{Journal of Scientific Computing},
vol.~92, no.~3,
Art.~no.~88,
2022.
\bibitem{Choi2026}
Y.-B. Choi and I.-B. Lee,
``Development of Physics-Informed Neural Networks (PINNs) for the natural ventilation of a greenhouse---Part 2: Development of prediction models for variable wind conditions,''
\emph{Biosystems Engineering},
vol.~265,
Art.~no.~104441,
2026.
\bibitem{Liu2025}
K. Liu, T. Ji, M. Li, X. Yang, J. Sun, H. Liu, and R. Liu,
``Prediction of greenhouse temperature and humidity across growing seasons: Hybridization of process-based model and deep neural networks,''
\emph{Information Processing in Agriculture},
vol.~13,
no.~2,
pp.~238--252,
2026. 


\bibitem{Salehi2024Dataset}
A. Salehi, A. Zeighmiyan, and A. Jafari,
``Greenhouse Shirvan Iran Dataset,''
\emph{Mendeley Data},
Version~2,
2024.
\url{https://doi.org/10.17632/ncrm9p4tc8.2}

\bibitem{Hull2023}
K. Hull, M. Mabitsela, E. Phiri, and M. Booysen,
``Dataset of temperature, humidity, and actuator states of an east-facing South African greenhouse tunnel,''
\emph{Data in Brief},
vol.~51,
Art.~no.~109633,
2023.

\bibitem{Wang2023DT}
Y. Wang, Z. Su, S. Guo, M. Dai, T. H. Luan, and Y. Liu,
``A survey on digital twins: Architecture, enabling technologies, security and privacy, and future prospects,''
\emph{IEEE Internet of Things Journal},
vol.~10,
no.~17,
pp.~14965--14987,
2023. 


\end{thebibliography}
\end{document}